\documentclass{article}
\usepackage{spconf,amsmath,graphicx,hyperref}
\usepackage{amssymb}  
\usepackage{multicol}
\usepackage{multirow}

\title{Mitigating Hallucination in Financial Retrieval-Augmented Generation via Fine-Grained Knowledge Verification}
\name{
\begin{tabular}{c}
Taoye Yin$^{\dagger}$\qquad Haoyuan Hu$^{\dagger}$ \qquad Yaxin Fan \qquad Xinhao Chen \\
Xinya Wu \qquad Kai Deng \qquad Kezun Zhang$^{\ddagger}$ \qquad Feng Wang
\end{tabular}
\thanks{$^{\dagger}$Co-first authors. $^{\ddagger}$Corresponding author: kezun.zkz@antgroup.com}
}

\address{
Ant Group, Hangzhou, China
}

\begin{document}

\maketitle
\begin{abstract}
In financial Retrieval-Augmented Generation (RAG) systems, models frequently rely on retrieved documents to generate accurate responses due to the time-sensitive nature of the financial domain. While retrieved documents help address knowledge gaps, model-generated responses still suffer from hallucinations that contradict the retrieved information. To mitigate this inconsistency, we propose a \textbf{R}einforcement \textbf{L}earning framework enhanced with \textbf{F}ine-grained \textbf{K}nowledge \textbf{V}erification (RLFKV).
Our method decomposes financial responses into atomic knowledge units and assesses the correctness of each unit to compute the fine-grained faithful reward. This reward offers more precise optimization signals, thereby improving alignment with the retrieved documents. Additionally, to prevent reward hacking (e.g., overly concise replies), we incorporate an informativeness reward that encourages the policy model to retain at least as many knowledge units as the base model.
Experiments conducted on the public Financial Data Description (FDD) task and our newly proposed FDD-ANT dataset demonstrate consistent improvements, confirming the effectiveness of our approach.
\end{abstract}
\begin{keywords}
Retrieval-augmented generation, Financial, Hallucination, Atomic knowledge
\end{keywords}

\section{Introduction}
\label{sec:intro}

With the increasing prevalence of large language models (LLMs), RAG has emerged as a critical technology for enhancing the accuracy and timeliness of LLM knowledge\cite{li2025searcho1agenticsearchenhancedlarge, zheng2025deepresearcherscalingdeepresearch, yu-etal-2024-chain}. By integrating LLMs with external knowledge bases, RAG effectively addresses key challenges such as knowledge obsolescence in LLMs\cite{NEURIPS2020_6b493230}.

\begin{figure}[t]
  \centering
  \includegraphics[scale=0.4]{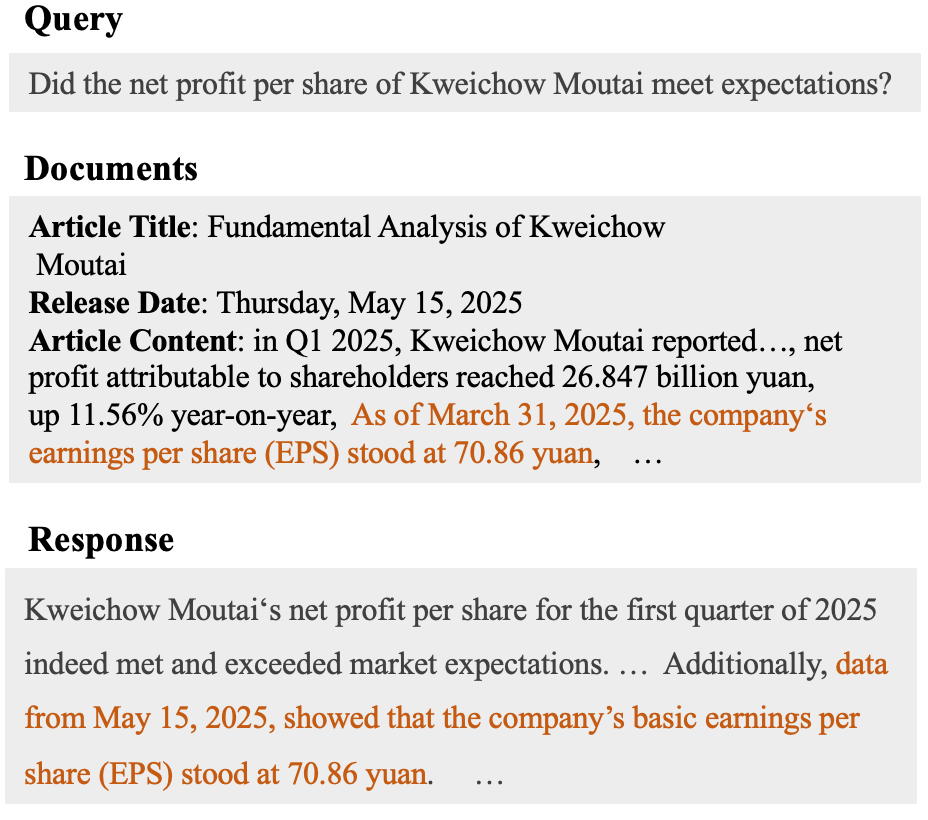}
  \caption{An example of hallucination by the base model Qwen3-8B in financial data description.} 
  \label{fig1} 
  \vspace{-15pt}
\end{figure}

\begin{figure*}[t]
\centering
\includegraphics[scale=0.5]{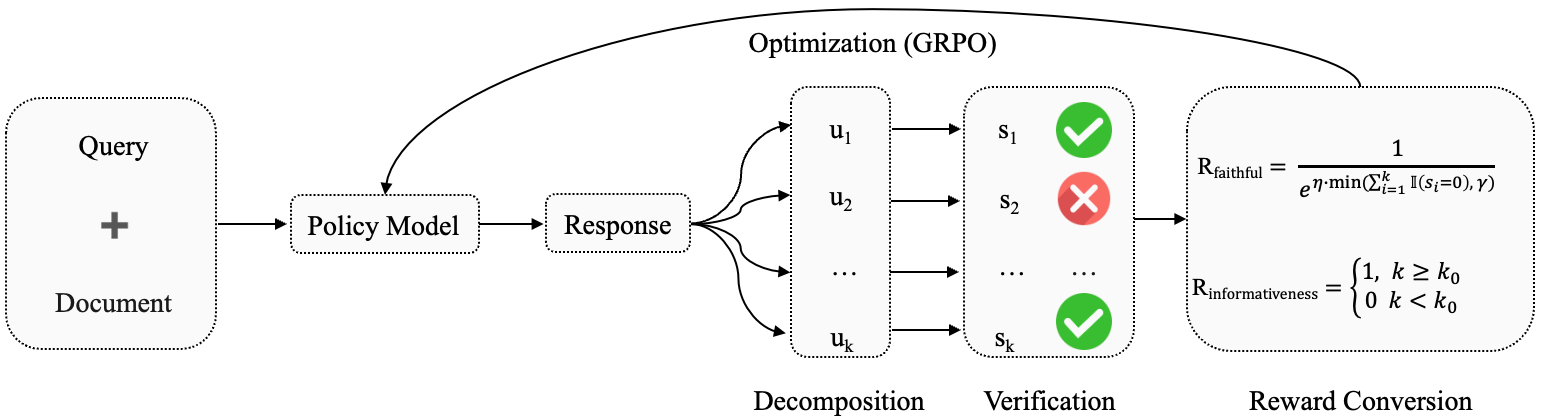} 
\caption{The framework of RLFKV.}
\label{fig2}
\vspace{-15pt}
\end{figure*}
In the financial domain, RAG systems always rely more heavily on retrieved documents due to the strong timeliness requirements of financial queries. As illustrated in Fig.~\ref{fig1}, for queries such as ``\emph{Did the net profit per share of Kweichow Moutai meet expectations?}'', the LLM must analyze the most recently retrieved documents to generate an accurate response. However, the generated answers may still contain hallucinations that conflict with the retrieved content. The orange highlighted section in Fig.~\ref{fig1} reveals a critical error: although the retrieved document explicitly states that as of March 31, 2025, the company's earnings per share (EPS) was 70.86 yuan, the model’s output incorrectly associates this value with May 15, 2025, demonstrating a clear temporal inconsistency.

Recent studies~\cite{guan2025deepragthinkingretrievestep, chen2025researchlearningreasonsearch, song2025r1searcherincentivizingsearchcapability, li2025r3raglearningstepbystepreasoning} have attempted to mitigate hallucinations in RAG systems using reinforcement learning, which typically relies on binary reward signals derived from human-annotated reference answers. However, these methods suffer from two major limitations: first, their reliance on manual answer annotation incurs high labeling costs; second, the discrete signals from coarse binary rewards fail to provide a stable optimization direction for model training.

To address these challenges, we propose the Reinforcement Learning with Fine-grained Knowledge Verification (RLFKV) framework. It enhances the factual consistency of generated responses with retrieved documents by providing fine-grained optimization signals, eliminating the need for annotated reference answers. Specifically, each response is decomposed into atomic knowledge units, which are minimal self-contained expressions of financial facts. These units are rigorously evaluated for their support in the retrieved documents, and the evaluation results provide fine-grained rewards to guide the model optimization. To prevent reward hacking whereby the model might generate overly brief responses, we introduce a binary pairwise constraint to ensure the policy model retains at least the same number of knowledge units as the base model. The policy model is optimized by maximizing the faithfulness and informativeness rewards jointly. Experiments on the public Financial Data Description (FDD) task from BizFinBench and our proposed FDD-ANT demonstrate the effectiveness of our method.

\section{Related work}
Previous research on RAG has primarily focused on improving LLM performance through document denoising \cite{fang-etal-2024-enhancing, hong-etal-2024-gullible}, query rewriting \cite{ma-etal-2023-query, zhang-etal-2024-adaptive}, and iterative retrieval \cite{yu2024autoragautonomousretrievalaugmentedgeneration, jiang-etal-2023-active}. The advancement of LLMs \cite{openai2024openaio1card, deepseekai2025deepseekr1incentivizingreasoningcapability} has further advanced RAG systems, especially through chain-of-thought reasoning for complex queries \cite{li2025searcho1agenticsearchenhancedlarge}.
 Recent studies \cite{guan2025deepragthinkingretrievestep, chen2025researchlearningreasonsearch, song2025r1searcherincentivizingsearchcapability, li2025r3raglearningstepbystepreasoning} have incorporated reinforcement learning to enhance LLMs, but these methods depend on reference answers and coarse-grained rewards. This approach not only incurs high costs for human annotation but also provides unstable guidance for optimizing response quality. In contrast, our RLFKV framework eliminates the need for annotated answers and provides fine-grained rewards for stable optimization, leading to higher-quality generation.

\section{METHOD}
Given the training query set $\mathcal{Q} = \{q_1, \dots, q_n\}$, retrieved documents $\mathcal{D} = \{d_1, \dots, d_m\}$, and the base model $\pi_0$ to be optimized, our goal is to learn a policy model $\pi$ that generates responses faithful to $\mathcal{D}$. As illustrated in Fig.~\ref{fig2}, our framework consists of two steps: 1) \textbf{Atomic Knowledge Unit Decomposition and Verification} (\textsection~\ref{section3_1}): The response $y_i$ is decomposed into atomic units using an evaluation model (Qwen3-32B), and each unit is then verified for consistency against $\mathcal{D}$. 2) \textbf{Optimization with Faithful and Informative Rewards} (\textsection~\ref{section3_2}). The evaluation results are converted into two reward signals: a faithful reward and an informative reward. Both these rewards collectively guide the optimization from $\pi_0$ to $\pi$, ensuring that the generated responses exhibit high faithfulness to the retrieved document $\mathcal{D}$ while retaining informativeness.
\subsection{Atomic Knowledge Unit Decomposition and Verification}
\label{section3_1}
For a query $q_i$ and its corresponding response $y_i$, the evaluation model first decomposes $y_i$ into a set of atomic knowledge units $U_i = \{u_1, u_2, \dots, u_{k}\}$, where each unit $u_j$ represents a minimal knowledge unit in the financial response. Then, the evaluation model verifies the factual consistency between each $u_j$ and the retrieved documents $D$, enabling the derivation of granular reward signals that guide model optimization.

\subsubsection{Atomic Knowledge Unit}
Building upon the atomicity of knowledge graph triplets, we introduce a financial quadruple structure (entity, metric, value, timestamp) to precisely capture the minimal knowledge units in financial-related descriptions. This design specifically addresses two core characteristics of financial texts: strict temporal sensitivity and quantitative-oriented representation. For example, the expression \emph{As of March 31, 2025, the company's earnings per share stood at 70.86 yuan} in Fig.~\ref{fig1} is represented as ``(Kweichow Moutai, basic earnings per share, 70.86 yuan, As of March 31, 2025)''. This quadruple structure enforces a completeness constraint where the absence of any element invalidates the factual assertion.

\subsubsection{Atomic Knowledge Unit Decomposition}
To decompose the response $y_i$ into atomic knowledge units, we designed a specialized prompt to guide the evaluation model. The decomposition process is formally defined as 
\begin{equation}
    \{u_i\}_{i=1}^k=f(y_i)
\end{equation}
where $u_i$ represents an atomic knowledge unit, $k$ denotes the number of atomic units in $y_i$, and $f$ is the evaluation model.
Our prompt engineering explicitly specifies four critical dimensions: entities, metrics, values, and timestamps. Specifically, we further provide a financial metric dictionary for the metric elements for reference.  Please refer to the code repository\footnote{\url{https://github.com/antgroup/ANT-Fin-RAG}} for prompt details.

\subsubsection{Atomic Knowledge Unit Verification}
Given the extracted atomic knowledge units $\{u_i\}_{i=1}^k$, we employ the evaluation model to assess their factual consistency with the retrieved document $\mathcal{D}$. This verification process is formally defined as:
\begin{equation}
    \{s_i\}_{i=1}^k = f(\{u_i\}_{i=1}^k, \mathcal{D})
\end{equation}
where $s_i \in \{0,1\}$ is a binary verification score,  and $1$ indicates $u_i$ is factually consistent with $\mathcal{D}$. 

\subsection{Optimization with Faithful and Informative Reward}
\label{section3_2}
After obtaining the fine-grained evaluation results $\{s_i\}_{i=1}^k$, we transform them into two distinct reward signals to guide model optimization: (1) a faithful reward $ r_{\text{f}}$ evaluating factual consistency, and (2) an informative reward $r_{\text{i}}$ assessing informational depth. $r_f$ is computed as:
\begin{equation}
    \begin{aligned}
        \text{score} &= \sum_{i=1}^k \mathbb{I}(s_i = 0), \\
        r_{\text{f}} &= \frac{1}{e^{\eta \cdot \min(\text{score}, \gamma)}}
    \end{aligned}
\end{equation}
where $\text{score}$ represents the count of incorrect atomic knowledge units, $\eta$ denotes the decay rate, min is the minimum function, and $\gamma$ serves as an upper limit for error counts. This formulation ensures the smooth reward by capping the penalty for excessive errors while maintaining sensitivity to factual inaccuracies.
$r_{\text{i}}$ is computed as:
\begin{equation}
r_{\text{i}} = \begin{cases}
1 & \text{if } k \geq k_0 \\ 
0 & \text{otherwise}
\end{cases}
\end{equation}
where $k_0$ represents the number of atomic knowledge units generated by the base model $\pi_0$. This binary reward scheme ensures that the policy model $\pi$ maintains at least the same level of informational depth as the base model, thereby effectively preventing reward hacking behavior. 

The overall reward $r$  is the average of the two rewards, and the GRPO\cite{shao2024deepseekmathpushinglimitsmathematical} is employed to optimize the model using the following objective function:
\begin{equation}
        \mathcal{L} = \mathbb{E}\left[ \sum_{i=1}^N \frac{\pi_\theta(a_i|s)}{\pi_{\theta_{\text{old}}}(a_i|s)} \cdot \frac{r_i - \bar{r}}{\sigma_r + \epsilon} - \beta \cdot \text{KL}(\pi_\theta \parallel \pi_{\theta_{\text{old}}}) \right]
\end{equation}
where $N$ is the number of samples in the group, $\pi_\theta$ is the optimized policy, $\pi_{\theta_{\text{old}}}$ is the old policy, $\bar{r}$ and  $\sigma_r$ are the mean and standard deviation of the rewards within the group, $\epsilon$ is a small constant for numerical stability, and the KL divergence term constrains the magnitude of the policy updates.

\section{Experiments}
\label{sec:experiment}

\subsection{Experiment Settings}
\textbf{Dataset.} We evaluated the effectiveness of the RLFKV framework on two datasets. The first is the Financial Data Description (FDD) dataset \cite{BizFinBench}, which is designed to assess a model’s ability to analyze financial data using retrieved documents. It consists of 1,461 samples but is limited to stock-related descriptions. To further validate the generalizability of our approach in real-world scenarios, we constructed a more diverse dataset, FDD-ANT, covering stocks, funds, and macroeconomic indicators. This dataset was collected from real business scenarios, with sensitive information manually redacted, and contains 2,000 samples. As neither dataset includes a training set, we followed the same protocol and randomly collected 4,000 samples for training purposes.

\noindent \textbf{Evaluation Metrics.}
We evaluate response quality along two dimensions: faithfulness and informativeness. For faithfulness (faith), we follow prior work \cite{BizFinBench} and employ a point-wise assessment using GPT-4o as the evaluator. Specifically, each response is classified into one of three categories: Fully Correct (100 points), Partially Correct (60 points), or Containing Significant Errors (0 points)\footnote{For detailed criteria, refer to \url{https://github.com/HiThink-Research/BizFinBench/blob/main/benchmark_code/BizFinBench/eval_financial_description.py}}. For informativeness (info), we use GPT-4o to count the number of atomic knowledge units present in each response and report the average across all samples.

\noindent \textbf{Baseline.}
We compare our model against the following baseline methods: general-purpose models including DeepSeek-V2-Lite-Chat-16B\cite{deepseekai2024deepseekv2strongeconomicalefficient}, Qwen3-8B\cite{yang2025qwen3technicalreport} and LLaMA3.1-8B-Instruct\cite{grattafiori2024llama3herdmodels}, as well as financial domain-specific models Xuanyuan-13B\cite{zhang2023xuanyuan20largechinese} and Dianjin-R1-7B\cite{zhu2025dianjinr1evaluatingenhancingfinancial}.

\noindent \textbf{Implementations.}
Our training process is implemented using the ms-swift framework. We employ Qwen3-8B and LLaMA3.1-8B-Instruct as base models and directly perform reinforcement learning. We train for 1 epoch with a learning rate of 1e-6. The batch size is set to 1 with 2 gradient accumulation steps. The maximum response length is 2048 tokens. The rollout number $N$ is set to 8. All experiments were conducted on 8 NVIDIA H20 GPUs.


\begin{table}[]
\centering

\begin{tabular}{lllll}
\hline
\multirow{2}{*}{Model}   & \multicolumn{2}{c}{FDD} & \multicolumn{2}{c}{FDD-ANT} \\
                         & faith     & info    & faith         & info         \\ \hline
                         \multicolumn{5}{l}{\emph{open-source model}} \\     
DeepSeek V2 Lite         & 69.1        &8.8             & 76.1              &  5.4                \\ 
LlaMA3            & 80.0          & 11.4            &   80.5            &   6.5              \\
Qwen3             &  86.5         & 13.4            &  90.2             &    10.8               \\ \hline
\multicolumn{5}{l}{\emph{finance model}}                                                    \\ 
Xuanyuan3                & 57.8         & 7.9            & 64.6              &5.8                 \\
Dianjin-R1               & 78.3         &  10.8           & 84.7              & 6.8                 \\ \hline
\multicolumn{5}{l}{\emph{ours}}                                                              \\ 
RLFKV$_{LLaMA3}$ & 83.6         & 11.7            & 82.1              &  8.1       \\
\quad w/o info reward            & 83.2          & 10.3         &  81.4               & 7.0                 \\ 
RLFKV$_{Qwen3}$ & \textbf{89.5 }         & \textbf{13.5}            &  \textbf{93.3}               & \textbf{12.3 }               \\ 
\quad w/o info reward   &  89.0         &  12.0           &  91.9            & 11.2               \\ \hline         
\end{tabular}
\caption{Performance comparison of various models on the FDD and FDD-ANT datasets.}
\label{tab:main_results}
\end{table}

\subsection{Main Result}
Table~\ref{tab:main_results} presents the performance comparison of various models on the FDD and FDD-ANT datasets. On the FDD dataset, our RLFKV\textsubscript{LLaMA3} improves the faithfulness score by 3.6 points over LLaMA3, while RLFKV\textsubscript{Qwen3} achieves a 3.0-point gain over Qwen3. A similar trend is observed on FDD-ANT, where RLFKV\textsubscript{LLaMA3} outperforms LLaMA3 by 1.6 points in faithfulness, and RLFKV\textsubscript{Qwen3} surpasses Qwen3 by 3.1 points. These results indicate that our method effectively reduces hallucination by generating responses that are more consistent with the retrieved documents.

When the informativeness reward is ablated (w/o info reward), the faithfulness score remains comparable, but the informativeness metric drops by 1.4 and 1.5 points for LLaMA3 and Qwen3, respectively, on FDD, and by 1.1 points for both on FDD-ANT. This suggests that optimizing solely for faithfulness may lead to shorter and less informative responses. In contrast, by integrating both rewards, our RLFKV framework enhances faithfulness while maintaining informativeness.

\subsection{Effectiveness of Fine-Grained Reward}
\begin{figure}[htb]

\begin{minipage}[b]{1.0\linewidth}
  \centering
  \centerline{\includegraphics[width=5cm]{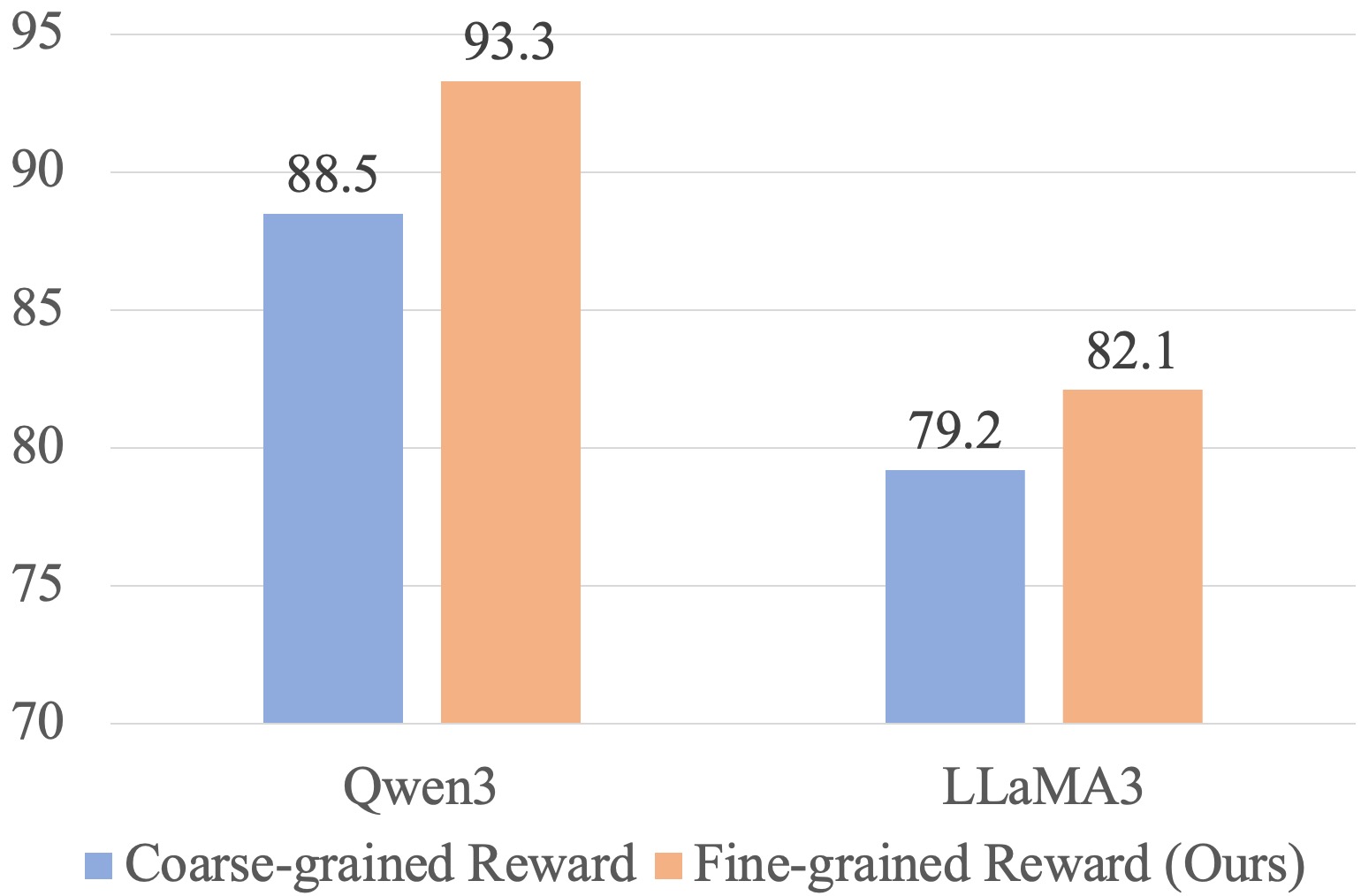}}
  \centerline{(a) Performance comparison on FDD-ANT datasets.}\medskip
\end{minipage}

\begin{minipage}[b]{1.0\linewidth}
  \centering
  \centerline{\includegraphics[width=5cm]{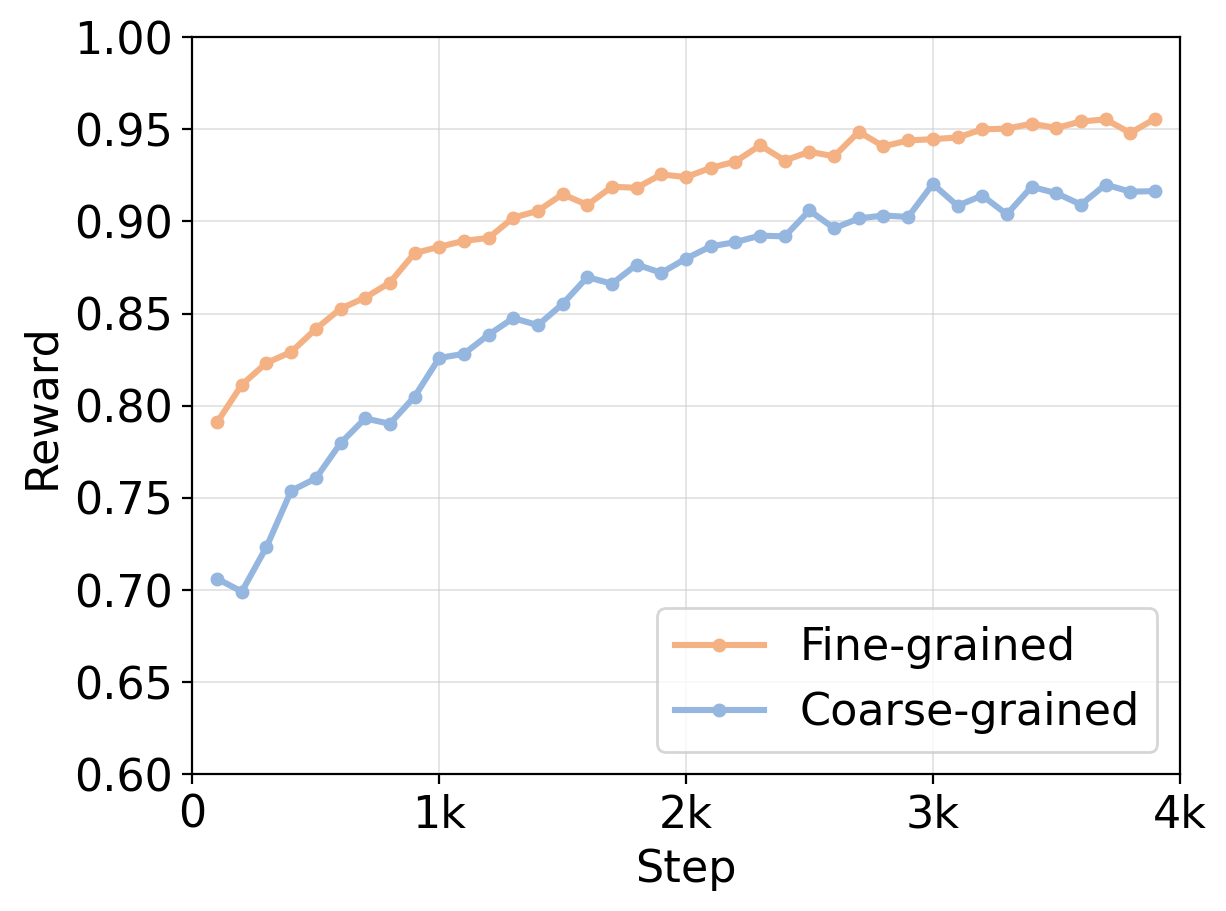}}
  \centerline{(b) Reward trends during the training process.}\medskip
\end{minipage}

\caption{Performance comparison and reward analysis of fine- vs. coarse-grained rewards based on Qwen3-8B model.}
\label{fig3}
\vspace{-5pt}
\end{figure}

To further demonstrate the effectiveness of our fine-grained reward, we compared it with the coarse-grained one that assigns binary rewards (1 if the response contains no factual errors, otherwise 0). As shown in Fig.~\ref{fig3}(a), models trained with the fine-grained reward achieved consistently higher faithful scores. Furthermore, Fig.~\ref{fig3}(b) illustrate the evolution of reward values during training. The fine-grained reward demonstrates more stable optimization and converges more rapidly (within 2k steps), indicating that it offers smoother learning signals and ultimately leads to superior performance.


\begin{figure}[t]
  \centering
  \includegraphics[scale=0.15]{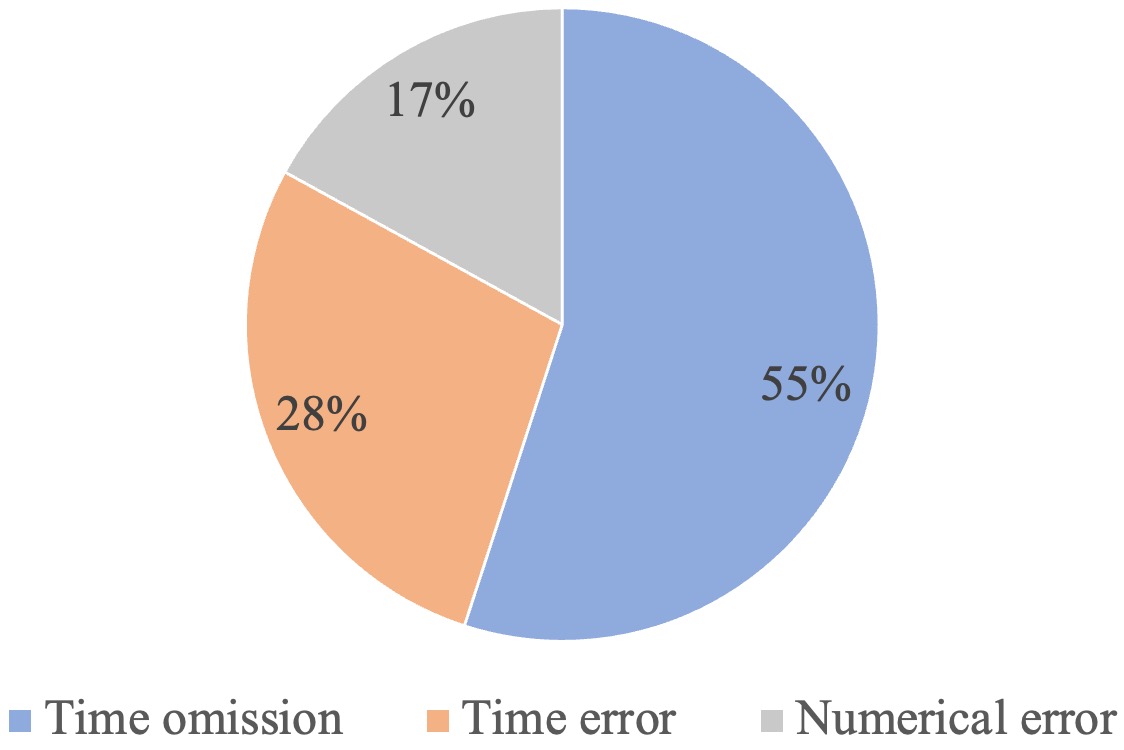} 
  \caption{Error analysis on FDD-ANT dataset.}
  \label{fig5}
  \vspace{-5pt}
\end{figure}
\subsection{Error Analysis}
We conducted an error analysis to further investigate potential improvements for our RLFKV framework. While our approach effectively reduces factual hallucinations, three persistent error types remain, as illustrated in Fig.~\ref{fig5}. The predominant issue is time omissions (55\%), where critical temporal references are absent despite being present in the retrieved documents. Other significant errors include time inaccuracies (28\%), particularly involving relative time expressions and fiscal-to-calendar year conversions, and numerical errors (17\%), which primarily occur in imprecise rounding. These findings illuminate promising pathways for future research and development.

\section{Conclusion}
In this paper, we propose a reinforcement learning framework with fine-grained knowledge verification (RLFKV) to improve the consistency between generated responses and retrieved documents. RLFKV decomposes responses into atomic knowledge units and evaluates the factuality of each unit, providing stable learning signals for model optimization. Additionally, we introduce FDD-ANT, a financial data description dataset spanning multiple data types, to validate the general applicability of our method in real-world scenarios. Experimental results on both the public FDD dataset and our released FDD-ANT dataset confirm the effectiveness of RLFKV. Future work will focus on refining reward mechanisms to enhance temporal and numerical accuracy.
\bibliographystyle{IEEEbib}
\bibliography{strings,refs}

\end{document}